\documentclass{article}
\usepackage{iclr2023_conference,times}

\usepackage{amsmath,amsfonts,bm}
\usepackage{svg}
\usepackage{hyperref}
\usepackage{url}
\usepackage{booktabs}
\usepackage{varwidth}
\usepackage{caption}
\usepackage[subtle]{savetrees}

\title{Mask Conditional Synthetic Satellite Imagery}

\author{Van Anh Le$^*$, Varshini Reddy\thanks{Equal Contribution}\ ,\ Zixi Chen$^*$, Mengyuan Li$^*$, Xinran Tang$^*$ \\
Institute for Applied Computational Science, Harvard University \\
\texttt{\{vananhle,varshinibogolu,zixichen,mengyuan\_li,xinran\_tang\}@g.harvard.edu}
\AND
Anthony Ortiz, Simone Fobi Nsutezo, Caleb Robinson \\
Microsoft AI for Good Research Lab \\
\texttt{\{anthony.ortiz,sfobinsutezo,caleb.robinson\}@microsoft.com}
}

\iclrfinalcopy
\begin{document}

\maketitle

\begin{abstract}
In this paper we propose a mask-conditional synthetic image generation model for creating synthetic satellite imagery datasets. Given a dataset of real high-resolution images and accompanying land cover masks, we show that it is possible to train an upstream conditional synthetic imagery generator, use that generator to create synthetic imagery with the land cover masks, then train a downstream model on the synthetic imagery and land cover masks that achieves similar test performance to a model that was trained with the real imagery. Further, we find that incorporating a mixture of real and synthetic imagery acts as a data augmentation method, producing better models than using only real imagery (0.5834 vs. 0.5235 mIoU). Finally, we find that encouraging diversity of outputs in the upstream model is a necessary component for improved downstream task performance. We have released code for reproducing our work on \href{https://github.com/ms-synthetic-satellite-image/synthetic-satellite-imagery}{GitHub}.
\end{abstract}

\section{Introduction}

Very high-resolution (VHR) satellite imagery offers unique insights into the state of the world, however using and releasing such imagery into the public domain often raises both privacy~\cite{coffer2020balancing} and legal questions. VHR satellite imagery can be used to rapidly assess the impacts of natural disasters~\citep{zheng2021building}, monitor urban growth~\citep{robinson2022fast}, and even re-identify individuals in populations of endangered whale species~\citep{hodul2023individual}, however it often cannot be shared publicly, which limits the ability of the research community to extend on previous work. In contrast, synthetic satellite imagery that is generated based on protected VHR imagery \textit{and that is useful for training models on downstream tasks} could be released with fewer concerns while simultaneously advancing application goals.

There has been work in computer vision literature focused on both the \textit{upstream} task of generating synthetic imagery, and various \textit{downstream} applications of synthetic imagery.
For the \textit{upstream task}, generative adversarial networks (GANs)~\cite{original_gan}, variational autoencoders (VAEs)~\cite{original_vae}, and denoising diffusion models~\cite{ho2020denoising} are all popular approaches for generating high-fidelity synthetic images. These models can be conditioned on labels or masks~\cite{cGAN,odena2017synthesize,zhu2020sean} to produce synthetic imagery guided by a desired class, semantic map, or image attributes. 

For \textit{downstream tasks}, including synthetic imagery in model training has shown to be: a strong data augmentation method for improving performance in liver lesion classification~\cite{frid2018synthetic}, method for balancing class distributions~\cite{wong2016understanding}, and as a domain adaptation method~\cite{deepsnow,kar2019meta}. Further, privacy preserving GANs train generators with differential privacy guarantees that can be used to create synthetic medical image datasets that can be shared without exposing personal data~\cite{liu2019ppgan,mukherjee2021privgan}.

Prior works have also focused on the specific domain of remote sensing and satellite imagery. For example SatGAN introduce a pixel-level reconstruction loss to generate colorful and blur-free versions of the ground truth satellite images~\cite{SatGan}, other work from the United Nations Satellite Centre employs a conditional progressive GAN to generate missing tiles in an image with applications to flood detection~\cite{cardoso2022conditional}, and ColorMapGAN propose training satellite images to match test images' spectral distributions to deal with the distribution shift~\cite{colormapgan}.

\begin{figure}
\includegraphics[width=1\linewidth]{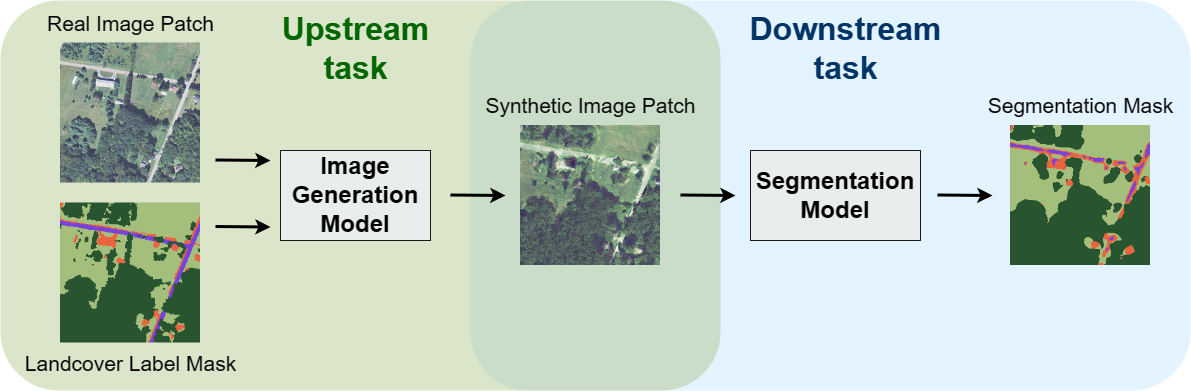}
\caption{\textbf{Modeling pipeline.} We first train an \textit{upstream model} that can generate synthetic imagery from label masks. Using this model we can create a dataset of synthetic imagery and accompanying label masks to train a \textit{downstream model}. We find that the downstream model trained on synthetic imagery achieves similar performance to one trained on real imagery on the land cover segmentation task.}
\label{fig:pipeline}
\end{figure}

In this work we combine ideas from both tasks to generate supervised synthetic satellite imagery data that can be used as a replacement for real data in downstream modeling tasks (Figure \ref{fig:pipeline}). Specifically, we train an upstream generator model on a real dataset of imagery and land cover masks, use this model to generate a dataset of synthetic imagery based on the real land cover masks, then test models trained on this synthetic dataset on held out data.
Here, the primary question is, ``\textbf{can a synthetic satellite imagery dataset provide similar utility as real dataset for modeling downstream tasks?}'' We find that 4-channel synthetic imagery is sufficient to train a downstream model to similar land cover performance as 3-channel real imagery (0.5171 vs. 0.5235 mIoU), and that synthetic imagery acts as a strong data augmentation method, improving downstream performance vs. real imagery alone by 6 points in mIoU. Our results suggest that conditional image generation is a promising approach for creating synthetic datasets of private satellite imagery that can be used by researchers to study downstream segmentation tasks.

\section{Methods}
Our work aims to generate synthetic satellite images which can substitute or augment real imagery used in training downstream models. Figure \ref{fig:pipeline} shows this pipeline, which consists of an upstream task -- synthetic data generation -- and a downstream task -- testing the usability of the synthetic data. We use the following method for the upstream model, and a simple U-Net~\cite{ronneberger2015u} for the downstream semantic segmentation task.

\paragraph{Mask-conditional image generation}
For the upstream model we employ a conditional GAN with spatially adaptive denormalization (SPADE)~\cite{spade} to generate mask-conditional synthetic satellite images. The model architecture consists of an encoder, a generator, and a discriminator. The encoder takes an input image and produces a mean and variance that is passed as an input to the generator. The generator uses a random vector $z$ from the reparameterized output of the encoder and the class mask to generate synthetic images through a series of SPADE ResNet blocks. Finally, the discriminator uses a PatchGAN~\cite{isola2017image} structure to evaluate the quality of the images generated. To increase diversity in the generator we add a maximization objective to the loss, following the idea from DSGANs~\cite{dsgan}: 

\begin{equation}
\mathcal{L}_{z}(G) = \lambda \;\; \mathbb{E}_{z_1,z_2} \left[ \min \left( \frac{\lVert G(x, z_1)-G(x,z_2) \rVert} {\lVert{z_{1}-z_{2}\rVert}}, \tau \right) \right] 
\end{equation}

Here $x$ is the label mask, $z$ is the latent embedding of an image from the generator, $\tau$ is a bound for numerical stability, and $\lambda$ controls the weights of the additional loss term and thus the diversity in the generated images. Intuitively, this loss term requires the generator to perform two forward passes and maximize the distance between the outputs.

\begin{figure}
\includegraphics[width=0.95\linewidth]{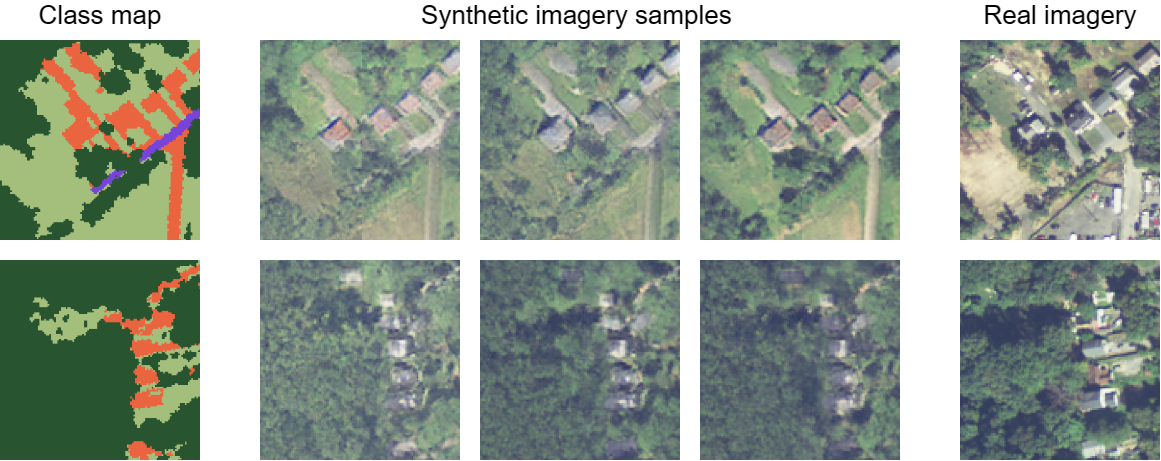}
\caption{Three examples of synthetic images generated from random latent representations for a given class mask from the $\lambda=6$ synthetic model. The ``real imagery'' is shown for reference but isn't used at inference time to generate the synthetic imagery.}
\label{fig:examples}
\end{figure}

\section{Experiments and Results}

\paragraph{Dataset}
We use the Maryland subset of the Chesapeake Land Cover dataset for our experiments. The dataset contains 4-channel high-resolution aerial imagery and high-resolution 6-class land cover labels from the Chesapeake Conservancy. The Maryland subset contains 100 $6000 \times 4000$ pixel train tiles, 5 validation tiles, and 20 test tiles~\cite{robinson2019large}\footnote{\url{https://lila.science/datasets/chesapeakelandcover}}.

\paragraph{Performance metrics.}
To evaluate the upstream model, we calculate the FID score over the test data. The FID score measures the distance between the distributions of Inception feature representation of real and synthetic images~\cite{heusel} and has been shown to correlate with human-perceived visual quality. A lower FID score suggests that our model can generate synthetic images that are like real images. 
To evaluate the performance of the downstream model, we compute average (mIoU) and label-wise intersection-over-union (IoU). 

\paragraph{Upstream Generation and Evaluation.}
\label{subsec:generate}
We trained 6 upstream models with different diversity levels by increasing the weight of the diversity loss term, $\lambda$, from $0$ to $10$ on a dataset of $256 \times 256$ RGB patches taken from 50 randomly selected train tiles. We train the models for four epochs with Xavier parameters initialization, a learning rate of $0.0002$ using the ADAM optimizer with $\beta_{1} = 0.0$ and $\beta_{2} = 0.9$, and a batch size of $10$. We additionally a single model with 4-channel inputs with $\lambda=0$ (to test the effect of input channels on performance).

After training, we generate 2 sets of synthetic images over the test tiles for upstream evaluation. In the first set, the generator takes a latent vector $z$ produced by the trained encoder. In the second set, $z$ is randomly sampled from a standard Gaussian. Our baseline model ($\lambda = 0$) achieves test FID scores of $73.7$ and $72.3$ for the first and second sets respectively (Table~\ref{tab:2}). For upstream model evaluation, using the trained encoder might not be appropriate if there are significant differences in the distributions of test tiles and training tiles. For both sets, models with higher values of $\lambda$ yield lower FID scores, suggesting that more diversity leads to synthetic images that are more similar to real images.

\begin{table}[th]
\begin{varwidth}[b]{0.45\linewidth}
\centering
\begin{tabular}{@{}cccc@{}}
\toprule
$\mathbf{\lambda}$ & \textbf{mIoU}   & $\textbf{FID}_a$  & $\textbf{FID}_b$   \\ \midrule
0                  & 0.2894          & 72.29         & 73.71    \\
2                  & 0.3417          & 63.07         & 70.72    \\
4                  & 0.3827          & 61.70         & \textbf{61.38}   \\
6                  & \textbf{0.4059} & \textbf{56.60}& 70.98    \\
8                  & 0.3572          & 58.09         & 63.46    \\
10                 & 0.3234          & 60.48         & 56.37    \\ \bottomrule
\end{tabular}
\caption{Impact of the $\lambda$ diversity parameter on the FID of generated images and downstream model performance. $\textbf{FID}_a$ is calculated with synthetic test images generated using random sampling from a standard Gaussian $\textbf{FID}_b$ is calculated with synthetic test images generated using the trained encoder.}
\label{tab:2}
\end{varwidth}%
\hfill
\begin{minipage}[b]{0.5\linewidth}
\centering
\includegraphics[width=0.95\linewidth]{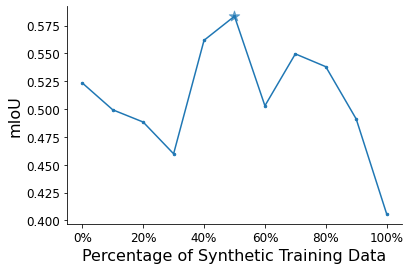}
\captionof{figure}{Impact of the proportion of synthetic images used when training a downstream model on segmentation performance (mIoU). Here, models are trained with $x\%$ of synthetic imagery and $100 - x\%$ of real imagery.}
\label{fig:mix}
\end{minipage}
\end{table}
\begin{table}[t]
\centering
\resizebox{\textwidth}{!}{%
\begin{tabular}{@{}lccccccc@{}}
\toprule
\multicolumn{1}{c}{\textbf{Method}} &
  \textbf{Water} &
  \textbf{Tree Canopy} &
  \textbf{Low Vegetation} &
  \textbf{Barren Land} &
  \textbf{\begin{tabular}[c]{@{}c@{}}Impervious\\ (other)\end{tabular}} &
  \textbf{\begin{tabular}[c]{@{}c@{}}Impervious\\ (road)\end{tabular}} &
  \textbf{Mean} \\ \midrule
100\% real      & 0.6794 & 0.8386 & 0.7279 & 0.1205 & 0.5302 & 0.2443 & 0.5235 \\
100\% synthetic & 0.4001 & 0.7332 & 0.5642 & 0.0134 & 0.4085 & 0.3161 & 0.4059 \\
200\% synthetic & 0.5322 & 0.6956 & 0.5636 & 0.0125 & 0.3677 & 0.3288 & 0.4167 \\
300\% synthetic & 0.2432 & 0.7402 & 0.5479 & 0.0157 & 0.3316 & 0.2878 & 0.3611 \\
100\% synthetic\textsuperscript{*} & 0.9100 & 0.7476 & 0.7034 & 0.0177 & 0.4143 & 0.3097 & 0.5171 \\ 
100\% real \textsuperscript{*}  & 0.9676 & 0.8532 & 0.8346 & 0.1456 & 0.5665 & 0.5137 & 0.6469
\\ \bottomrule
\end{tabular}
}
\caption{Performance (IoU) of downstream models trained on real and synthetic imagery measured on the Maryland test set. 100\% synthetic, 200\% synthetic, and 300\% synthetic are generated with the diversity parameter $\lambda=6$. \textsuperscript{*} means the imagery is 4-channel. The 100\% synthetic\textsuperscript{*} result is from prior experiments without the diversity improvement loss, and only uses 2000 $256\times256$ pixel patches.}
\label{tab:tab1}
\end{table}
\paragraph{Downstream Evaluation.} \label{subsec:downstream}
We test whether our synthetic satellite imagery can be effectively used for data substitution or augmentation with a series of downstream experiments using the land cover mapping task. In each experiment we evaluate the performance of the downstream models on the held out test set from the Maryland split of the dataset.

We first explore the downstream performance of models trained with real label masks and synthetic imagery generated with different degrees of diversity, i.e. from upstream models trained with different values of $\lambda$. As shown in Table~\ref{tab:2}, synthetic imagery tiles generated with $\lambda=6$ result in the highest mIoU scores of 0.4059. As such, we use tiles generated with the upstream model trained with $\lambda=6$ for most of the following experiments.

To evaluate the usability of the synthetic images with respect to the real ones, we trained four U-Nets with a learning rate of $1e-4$, an early stopping patience of 10 and otherwise matching hyperparameters to the upstream models. We use the following datasets: 100\% real, which contains 100 real satellite image tiles; 100\% synthetic, 200\% synthetic, 300\% synthetic, which contains 100, 200, 300 synthetic RGB tiles respectively. For the datasets $>100\%$ we sample multiple synthetic images for each land cover mask with different randomly encoded latent vectors. Figure~\ref{fig:examples} shows examples of these synthetic image patches.

Table~\ref{tab:tab1} shows that the models trained only on synthetic imagery have worse performance, mainly caused by decreased performance in the water and barren classes. The NIR spectral channel contains useful information for water identification~\cite{yuan2021deep}, leading to more accurate segmentation results, so we trained two other models on 100\% real\textsuperscript{*} and 100\% synthetic\textsuperscript{*} separately with the full 4-channel RGB-NIR information in the dataset\footnote{The 100\% synthetic\textsuperscript{*} model is trained on 2000 $256\times256$ 4-channel patches generated from the upstream model with baseline diversity $\lambda=0$, while the other methods all randomly sampled 200 $256\times256$ patches per tile, thus 20000 patches in total when training.}. Table \ref{tab:tab1} shows that, even with only 10\% the amount of data, the segmentation model trained on 4-channel synthetic images could reach comparable performance as the model trained on 3-channel real images. The models trained on real 4-channel imagery perform better than all others.

Finally, we test whether synthetic imagery can be used to augment real imagery in model training. Let $p$ be the proportion of synthetic images used in training. We create datasets by randomly selecting $100p$ synthetic tiles generated from the upstream model using $\lambda=6$ and the corresponding $100(1-p)$ remaining original real tiles. We train downstream model on datasets with different values of $p$ from 0\% to 100\%. Figure~\ref{fig:mix} shows the results of this experiment and we observe that the model trained with $p=50\%$ achieves a higher mIoU, 0.5834, than the mIoU of a model trained only with real images, 0.5235.

\section{Conclusion}

We show that it is possible to generate a synthetic satellite imagery dataset with a mask-conditional GAN that can be used as a substitute for real imagery in a downstream land cover mapping task. This method can potentially be used to generate synthetic versions of protected VHR imagery datasets that could be released publicly for further study. Future work down this line of research include: studying how to generate large amounts of synthetic satellite imagery without tiling artifacts, evaluating the utility of synthetic imagery datasets in a variety of downstream tasks, investigating whether training data can be recovered from the generator model (i.e. whether privacy preserving methods are necessary), and testing different methods for upstream image generation.

\bibliography{citations}

\begin{thebibliography}{26}
\providecommand{\natexlab}[1]{#1}
\providecommand{\url}[1]{\texttt{#1}}
\expandafter\ifx\csname urlstyle\endcsname\relax
  \providecommand{\doi}[1]{doi: #1}\else
  \providecommand{\doi}{doi: \begingroup \urlstyle{rm}\Url}\fi

\bibitem[Cardoso et~al.(2022)Cardoso, Vallecorsa, and
  Nemni]{cardoso2022conditional}
Renato Cardoso, Sofia Vallecorsa, and Edoardo Nemni.
\newblock Conditional progressive generative adversarial network for satellite
  image generation.
\newblock \emph{arXiv preprint arXiv:2211.15303}, 2022.

\bibitem[Coffer(2020)]{coffer2020balancing}
Megan~M Coffer.
\newblock Balancing privacy rights and the production of high-quality satellite
  imagery, 2020.

\bibitem[Frid-Adar et~al.(2018)Frid-Adar, Klang, Amitai, Goldberger, and
  Greenspan]{frid2018synthetic}
Maayan Frid-Adar, Eyal Klang, Michal Amitai, Jacob Goldberger, and Hayit
  Greenspan.
\newblock Synthetic data augmentation using gan for improved liver lesion
  classification.
\newblock In \emph{2018 IEEE 15th international symposium on biomedical imaging
  (ISBI 2018)}, pp.\  289--293. IEEE, 2018.

\bibitem[Ghahramani et~al.(2014)Ghahramani, Welling, Cortes, Lawrence, and
  Weinberger]{original_gan}
Z~Ghahramani, M~Welling, C~Cortes, ND~Lawrence, and KQ~Weinberger.
\newblock Generative adversarial nets.
\newblock \emph{Advances in Neural Information Processing Systems},
  27:\penalty0 2672--2680, 2014.

\bibitem[Heusel et~al.(2017)Heusel, Ramsauer, Unterthiner, Nessler, and
  Hochreiter]{heusel}
Martin Heusel, Hubert Ramsauer, Thomas Unterthiner, Bernhard Nessler, and Sepp
  Hochreiter.
\newblock Gans trained by a two time-scale update rule converge to a local nash
  equilibrium.
\newblock \emph{Advances in neural information processing systems}, 30, 2017.

\bibitem[Ho et~al.(2020)Ho, Jain, and Abbeel]{ho2020denoising}
Jonathan Ho, Ajay Jain, and Pieter Abbeel.
\newblock Denoising diffusion probabilistic models.
\newblock \emph{Advances in Neural Information Processing Systems},
  33:\penalty0 6840--6851, 2020.

\bibitem[Hodul et~al.(2023)Hodul, Knudby, McKenna, James, Mayo, Brown,
  Durette-Morin, and Bird]{hodul2023individual}
Matus Hodul, Anders Knudby, Brigid McKenna, Amy James, Charles Mayo, Moira
  Brown, Delphine Durette-Morin, and Stephen Bird.
\newblock Individual north atlantic right whales identified from space.
\newblock \emph{Marine Mammal Science}, 39\penalty0 (1):\penalty0 220--231,
  2023.

\bibitem[Isola et~al.(2017)Isola, Zhu, Zhou, and Efros]{isola2017image}
Phillip Isola, Jun-Yan Zhu, Tinghui Zhou, and Alexei~A Efros.
\newblock Image-to-image translation with conditional adversarial networks.
\newblock In \emph{Proceedings of the IEEE conference on computer vision and
  pattern recognition}, pp.\  1125--1134, 2017.

\bibitem[Kar et~al.(2019)Kar, Prakash, Liu, Cameracci, Yuan, Rusiniak, Acuna,
  Torralba, and Fidler]{kar2019meta}
Amlan Kar, Aayush Prakash, Ming-Yu Liu, Eric Cameracci, Justin Yuan, Matt
  Rusiniak, David Acuna, Antonio Torralba, and Sanja Fidler.
\newblock Meta-sim: Learning to generate synthetic datasets.
\newblock In \emph{Proceedings of the IEEE/CVF International Conference on
  Computer Vision}, pp.\  4551--4560, 2019.

\bibitem[Kingma \& Welling(2013)Kingma and Welling]{original_vae}
Diederik~P Kingma and Max Welling.
\newblock Auto-encoding variational bayes.
\newblock \emph{arXiv preprint arXiv:1312.6114}, 2013.

\bibitem[Liu et~al.(2019)Liu, Peng, James, and Wu]{liu2019ppgan}
Yi~Liu, Jialiang Peng, JQ~James, and Yi~Wu.
\newblock Ppgan: Privacy-preserving generative adversarial network.
\newblock In \emph{2019 IEEE 25Th international conference on parallel and
  distributed systems (ICPADS)}, pp.\  985--989. IEEE, 2019.

\bibitem[Mirza \& Osindero(2014)Mirza and Osindero]{cGAN}
Mehdi Mirza and Simon Osindero.
\newblock Conditional generative adversarial nets.
\newblock \emph{arXiv preprint arXiv:1411.1784}, 2014.

\bibitem[Mukherjee et~al.(2021)Mukherjee, Xu, Trivedi, Patowary, and
  Ferres]{mukherjee2021privgan}
Sumit Mukherjee, Yixi Xu, Anusua Trivedi, Nabajyoti Patowary, and Juan~Lavista
  Ferres.
\newblock privgan: Protecting gans from membership inference attacks at low
  cost to utility.
\newblock \emph{Proc. Priv. Enhancing Technol.}, 2021\penalty0 (3):\penalty0
  142--163, 2021.

\bibitem[Odena et~al.(2017)Odena, Olah, and Shlens]{odena2017synthesize}
Augustus Odena, Christopher Olah, and Jonathon Shlens.
\newblock Conditional image synthesis with auxiliary classifier gans.
\newblock \emph{Web: https://arxiv.org/pdf/1610.09585.pdf}, 2017.

\bibitem[Park et~al.(2019)Park, Liu, Wang, and Zhu]{spade}
Taesung Park, Ming-Yu Liu, Ting-Chun Wang, and Jun-Yan Zhu.
\newblock Semantic image synthesis with spatially-adaptive normalization.
\newblock In \emph{Proceedings of the IEEE/CVF conference on computer vision
  and pattern recognition}, pp.\  2337--2346, 2019.

\bibitem[Ren et~al.(2020)Ren, Ziemann, Theiler, and Durieux]{deepsnow}
Christopher~X Ren, Amanda Ziemann, James Theiler, and Alice~MS Durieux.
\newblock Deep snow: synthesizing remote sensing imagery with generative
  adversarial nets.
\newblock In \emph{Algorithms, Technologies, and Applications for Multispectral
  and Hyperspectral Imagery XXVI}, volume 11392, pp.\  196--205. SPIE, 2020.

\bibitem[Robinson et~al.(2019)Robinson, Hou, Malkin, Soobitsky, Czawlytko,
  Dilkina, and Jojic]{robinson2019large}
Caleb Robinson, Le~Hou, Kolya Malkin, Rachel Soobitsky, Jacob Czawlytko, Bistra
  Dilkina, and Nebojsa Jojic.
\newblock Large scale high-resolution land cover mapping with multi-resolution
  data.
\newblock In \emph{Proceedings of the IEEE Conference on Computer Vision and
  Pattern Recognition}, pp.\  12726--12735, 2019.

\bibitem[Robinson et~al.(2022)Robinson, Ortiz, Park, Lozano, Kaw, Sederholm,
  Dodhia, and Ferres]{robinson2022fast}
Caleb Robinson, Anthony Ortiz, Hogeun Park, Nancy Lozano, Jon~Kher Kaw, Tina
  Sederholm, Rahul Dodhia, and Juan M~Lavista Ferres.
\newblock Fast building segmentation from satellite imagery and few local
  labels.
\newblock In \emph{Proceedings of the IEEE/CVF Conference on Computer Vision
  and Pattern Recognition}, pp.\  1463--1471, 2022.

\bibitem[Ronneberger et~al.(2015)Ronneberger, Fischer, and
  Brox]{ronneberger2015u}
Olaf Ronneberger, Philipp Fischer, and Thomas Brox.
\newblock U-net: Convolutional networks for biomedical image segmentation.
\newblock In \emph{Medical Image Computing and Computer-Assisted
  Intervention--MICCAI 2015: 18th International Conference, Munich, Germany,
  October 5-9, 2015, Proceedings, Part III 18}, pp.\  234--241. Springer, 2015.

\bibitem[Shah et~al.(2021)Shah, Gupta, and Thakkar]{SatGan}
Mitt Shah, Manish Gupta, and Priyank Thakkar.
\newblock Satgan: Satellite image generation using conditional adversarial
  networks.
\newblock In \emph{2021 International Conference on Communication information
  and Computing Technology (ICCICT)}, pp.\  1--6. IEEE, 2021.

\bibitem[Tasar et~al.(2020)Tasar, Happy, Tarabalka, and Alliez]{colormapgan}
Onur Tasar, SL~Happy, Yuliya Tarabalka, and Pierre Alliez.
\newblock Colormapgan: Unsupervised domain adaptation for semantic segmentation
  using color mapping generative adversarial networks.
\newblock \emph{IEEE Transactions on Geoscience and Remote Sensing},
  58\penalty0 (10):\penalty0 7178--7193, 2020.

\bibitem[Wong et~al.(2016)Wong, Gatt, Stamatescu, and
  McDonnell]{wong2016understanding}
Sebastien~C Wong, Adam Gatt, Victor Stamatescu, and Mark~D McDonnell.
\newblock Understanding data augmentation for classification: when to warp?
\newblock In \emph{2016 international conference on digital image computing:
  techniques and applications (DICTA)}, pp.\  1--6. IEEE, 2016.

\bibitem[Yang et~al.(2019)Yang, Hong, Jang, Zhao, and Lee]{dsgan}
Dingdong Yang, Seunghoon Hong, Yunseok Jang, Tianchen Zhao, and Honglak Lee.
\newblock Diversity-sensitive conditional generative adversarial networks.
\newblock \emph{arXiv preprint arXiv:1901.09024}, 2019.

\bibitem[Yuan et~al.(2021)Yuan, Zhuang, Schaefer, Feng, Guan, and
  Fang]{yuan2021deep}
Kunhao Yuan, Xu~Zhuang, Gerald Schaefer, Jianxin Feng, Lin Guan, and Hui Fang.
\newblock Deep-learning-based multispectral satellite image segmentation for
  water body detection.
\newblock \emph{IEEE Journal of Selected Topics in Applied Earth Observations
  and Remote Sensing}, 14:\penalty0 7422--7434, 2021.

\bibitem[Zheng et~al.(2021)Zheng, Zhong, Wang, Ma, and
  Zhang]{zheng2021building}
Zhuo Zheng, Yanfei Zhong, Junjue Wang, Ailong Ma, and Liangpei Zhang.
\newblock Building damage assessment for rapid disaster response with a deep
  object-based semantic change detection framework: From natural disasters to
  man-made disasters.
\newblock \emph{Remote Sensing of Environment}, 265:\penalty0 112636, 2021.

\bibitem[Zhu et~al.(2020)Zhu, Abdal, Qin, and Wonka]{zhu2020sean}
Peihao Zhu, Rameen Abdal, Yipeng Qin, and Peter Wonka.
\newblock Sean: Image synthesis with semantic region-adaptive normalization.
\newblock In \emph{Proceedings of the IEEE/CVF Conference on Computer Vision
  and Pattern Recognition}, pp.\  5104--5113, 2020.

\end{thebibliography}
\bibliographystyle{iclr2023_conference}

\end{document}